%% file: conll2018.tex
\documentclass[11pt,a4paper]{article}
\usepackage[hyperref]{conll2018}
\usepackage{times}
\usepackage{latexsym}
\usepackage{multirow}
\usepackage{graphicx}
\usepackage{lipsum}
\usepackage{courier}
\usepackage{enumitem}
\usepackage{amsmath}
\usepackage{color}
\usepackage{subcaption}
\usepackage{tabularx}
\usepackage{CJKutf8}
\usepackage{booktabs}
\usepackage{mathrsfs}
\usepackage{xcolor}
\usepackage[
  separate-uncertainty = true,
  multi-part-units = repeat
]{siunitx}
\usepackage{hhline}
\usepackage{url}

\aclfinalcopy 

\title{Global Attention for Name Tagging}

\author{Boliang Zhang, Spencer Whitehead, Lifu Huang and Heng Ji \\
	    Computer Science Department \\
        Rensselaer Polytechnic Institute\\
        {\tt \{zhangb8,whites5,huangl7,jih\}@rpi.edu}\\
        }

\definecolor{good_entity}{RGB}{0,153,76}
\definecolor{bad_entity}{RGB}{255,51,51}

\date{}

\begin{document}
\begin{CJK*}{UTF8}{gbsn}
\maketitle

\input{0abstract}
\input{1introduction}
\input{3model}
\input{4experiments}
\input{2related}
\input{5conclusions}
\section*{Acknowledgments}
This work was supported by the U.S. DARPA AIDA Program No. FA8750-18-2-0014, LORELEI Program No. HR0011-15-C-0115, Air Force No. FA8650-17-C-7715, NSF IIS-1523198 and U.S. ARL NS-CTA No. W911NF-09-2-0053. The views and conclusions contained in this document are those of the authors and should not be interpreted as representing the official policies, either expressed or implied, of the U.S. Government. The U.S. Government is authorized to reproduce and distribute reprints for Government purposes notwithstanding any copyright notation here on.

\bibliography{conll2018}
\bibliographystyle{acl_natbib_nourl}

\clearpage\end{CJK*}
\end{document}

%% file: 0abstract.tex
\begin{abstract}
Many name tagging approaches use local contextual information with much success, but fail when the local context is ambiguous or limited. We present a new framework to improve name tagging by utilizing local, document-level, and corpus-level contextual information. We retrieve document-level context from other sentences within the same document and corpus-level context from sentences in other topically related documents. We propose a model that learns to incorporate document-level and corpus-level contextual information alongside local contextual information via global 
attentions, which dynamically weight their respective contextual information, and gating mechanisms, which determine the influence of this information. Extensive experiments on benchmark datasets show the effectiveness of our approach, which achieves state-of-the-art results for Dutch, German, and Spanish on the CoNLL-2002 and CoNLL-2003 datasets.\footnote{The programs are publicly available for research purpose: \url{https://github.com/boliangz/global_attention_ner}}.

\end{abstract}


%% file: 1introduction.tex
\section{Introduction}

\begin{figure}[h!t]
\centering\small
\includegraphics[width=0.48\textwidth]{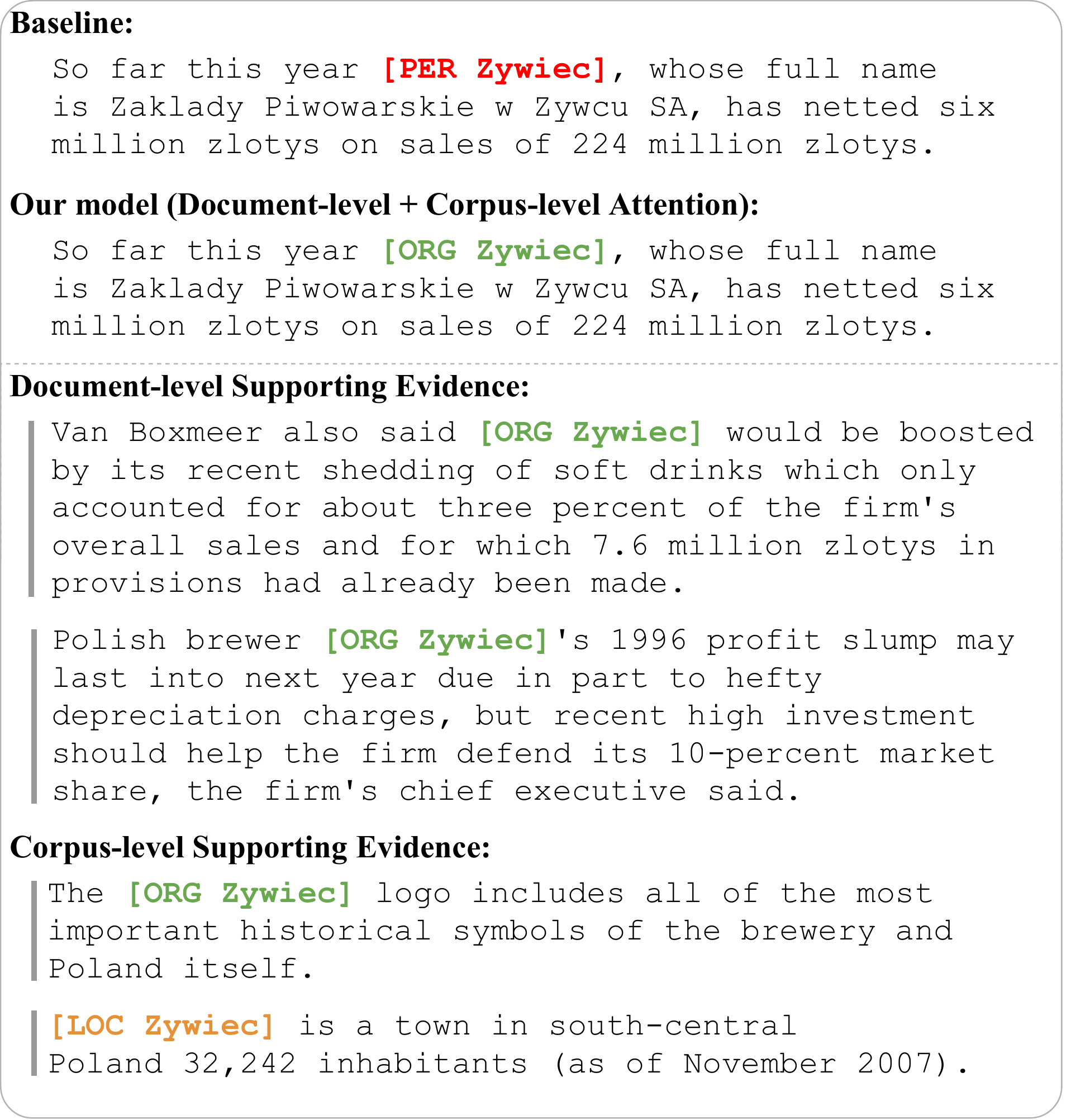}
   \caption{Example from the baseline and our model with some supporting evidence.}
\vspace{+0.5em}  
\label{fig:global_context_example}
\end{figure}

Name tagging, the task of automatically identifying and classifying named entities in text, is often posed as a sentence-level sequence labeling problem where each token is labeled as being part of a name of a certain type ({\sl e.g.,} location) or not~\cite{chinchor1997muc,tjong2003introduction}. When labeling a token, local context ({\sl i.e.,} surrounding tokens) is crucial because the context gives insight to the semantic meaning of the token. However, there are many instances in which the local context is ambiguous or lacks sufficient content. For example, in Figure~\ref{fig:global_context_example}, the query sentence discusses ``\texttt{Zywiec}'' selling a product and profiting from these sales, but the local contextual information is ambiguous as more than one entity type could be involved in a sale. As a result, the baseline model mistakenly tags ``\texttt{Zywiec}'' as a person (PER) instead of the correct tag, which is organization (ORG). If the model has access to supporting evidence that provides additional, clearer contextual information, then the model may use this information to correct the mistake given the ambiguous local context.
\graphicspath { {./fig/} }

Additional context may be found from other sentences in the same document as the query sentence (\textbf{document-level}). In Figure~\ref{fig:global_context_example}, the sentences in the document-level supporting evidence provide clearer clues to tag ``\texttt{Zywiec}'' as ORG, such as the references to ``\texttt{Zywiec}'' as a ``\texttt{firm}''. A concern of leveraging this information is the amount of noise that is introduced. However, across all the data in our experiments (Section~\ref{subsec:data}), we find that an average of 35.43\% of named entity mentions in each document are repeats and, when a mention appears more than once in a document, an average of 98.78\% of these mentions have the same type. Consequently, one may use the document-level context to overcome the ambiguities of the local context while introducing little noise.

Although a significant amount of named entity mentions are repeated, 64.57\% of the mentions are unique. In such cases, the sentences at the document-level cannot serve as a source of additional context. Nevertheless, one may find additional context from sentences in other documents in the corpus (\textbf{corpus-level}). Figure~\ref{fig:global_context_example} shows some of the corpus-level supporting evidence for ``\texttt{Zywiec}''. In this example, similar to the document-level supporting evidence, the first sentence in this corpus-level evidence discusses the branding of ``\texttt{Zywiec}'', corroborating the ORG tag. Whereas the second sentence introduces noise because it has a different topic than the current sentence and discusses the Polish town named ``\texttt{Zywiec}'', one may filter these noisy contexts, especially when the noisy contexts are accompanied by clear contexts like the first sentence.

We propose to utilize local, document-level, and corpus-level contextual information to improve name tagging. Generally, we follow the \textit{one sense per discourse} hypothesis introduced by~\citet{Yarowsky1995}. Some previous name tagging efforts apply this hypothesis to conduct majority voting for multiple mentions with the same name string in a discourse through a cache model~\cite{Florian2004} or post-processing~\cite{Hermjakob2017}. However, these rule-based methods require manual tuning of thresholds. Moreover, it's challenging to explicitly define the scope of discourse. We propose a new neural network framework with global attention to tackle these challenges. Specifically, for each token in a query sentence, we propose to retrieve sentences that contain the same token from the document-level and corpus-level contexts ({\sl e.g.,} document-level and corpus-level supporting evidence for ``\texttt{Zywiec}'' in Figure~\ref{fig:global_context_example}). To utilize this additional information, we propose a model that, first, produces representations for each token that encode the local context from the query sentence as well as the document-level and corpus-level contexts from the retrieved sentences. Our model uses a \textit{document-level attention} and \textit{corpus-level attention} to dynamically weight the document-level and corpus-level contextual representations, emphasizing the contextual information from each level that is most relevant to the local context and filtering noise such as the irrelevant information from the mention ``\texttt{[LOC Zywiec]}'' in Figure~\ref{fig:global_context_example}. The model learns to balance the influence of the local, document-level, and corpus-level contextual representations via gating mechanisms. Our model predicts a tag using the local, gated-attentive document-level, and gated-attentive corpus-level contextual representations, which allows our model to predict the correct tag, ORG, for ``\texttt{Zywiec}'' in Figure~\ref{fig:global_context_example}.

The major contributions of this paper are: First, we propose to use multiple levels of contextual information (local, document-level, and corpus-level) to improve name tagging. Second, we present two new attentions, document-level and corpus-level, which prove to be effective at exploiting extra contextual information and achieve the state-of-the-art.

%% file: 3model.tex
\section{Model}
We first introduce our baseline model. Then, we enhance this baseline model by adding document-level and corpus-level contextual information to the prediction process via our document-level and corpus-level attention mechanisms, respectively. 

\input{3.1baseline}

\input{3.2doc_attention}

\input{3.3corpus_attention}
\input{3.4prediction}

%% file: 3.1baseline.tex
\subsection{Baseline}

We consider name tagging as a sequence labeling problem, where each token in a sequence is tagged as the beginning (B), inside (I) or outside (O) of a name mention. The tagged names are then classified into predefined entity types. In this paper, we only use the person (PER), organization (ORG), location (LOC), and miscellaneous (MISC) types, which are the predefined types in CoNLL-02 and CoNLL-03 name tagging dataset~\cite{tjong2003introduction}.

Our baseline model has two parts: 1) Encoding the sequence of tokens by incorporating the preceding and following contexts using a bi-directional long short-term memory (Bi-LSTM)~\cite{graves2013hybrid}, so each token is assigned a local contextual embedding. Here, following~\citet{ma2016end}, we use the concatenation of pre-trained word embeddings and character-level word representations composed by a convolutional neural network (CNN) as input to the Bi-LSTM. 2) Using a Conditional Random Fields (CRFs) output layer to render predictions for each token, which can efficiently capture dependencies among name tags ({\sl e.g.,} ``I-LOC'' cannot follow ``B-ORG'').

The Bi-LSTM CRF network is a strong baseline due to its remarkable capability of modeling contextual information and label dependencies. Many recent efforts combine the Bi-LSTM CRF network with language modeling~\cite{liu2017empower,peters2017semi,peters2018deep} to boost the name tagging performance. However, they still suffer from the limited contexts within individual sequences. To overcome this limitation, we introduce two attention mechanisms to incorporate document-level and corpus-level supporting evidence.

%% file: 3.2doc_attention.tex
\subsection{Document-level Attention}
\label{section:doc-lvl}

\graphicspath { {./fig/} }
\begin{figure*}[h!t]
\centering
\includegraphics[width=1\textwidth]{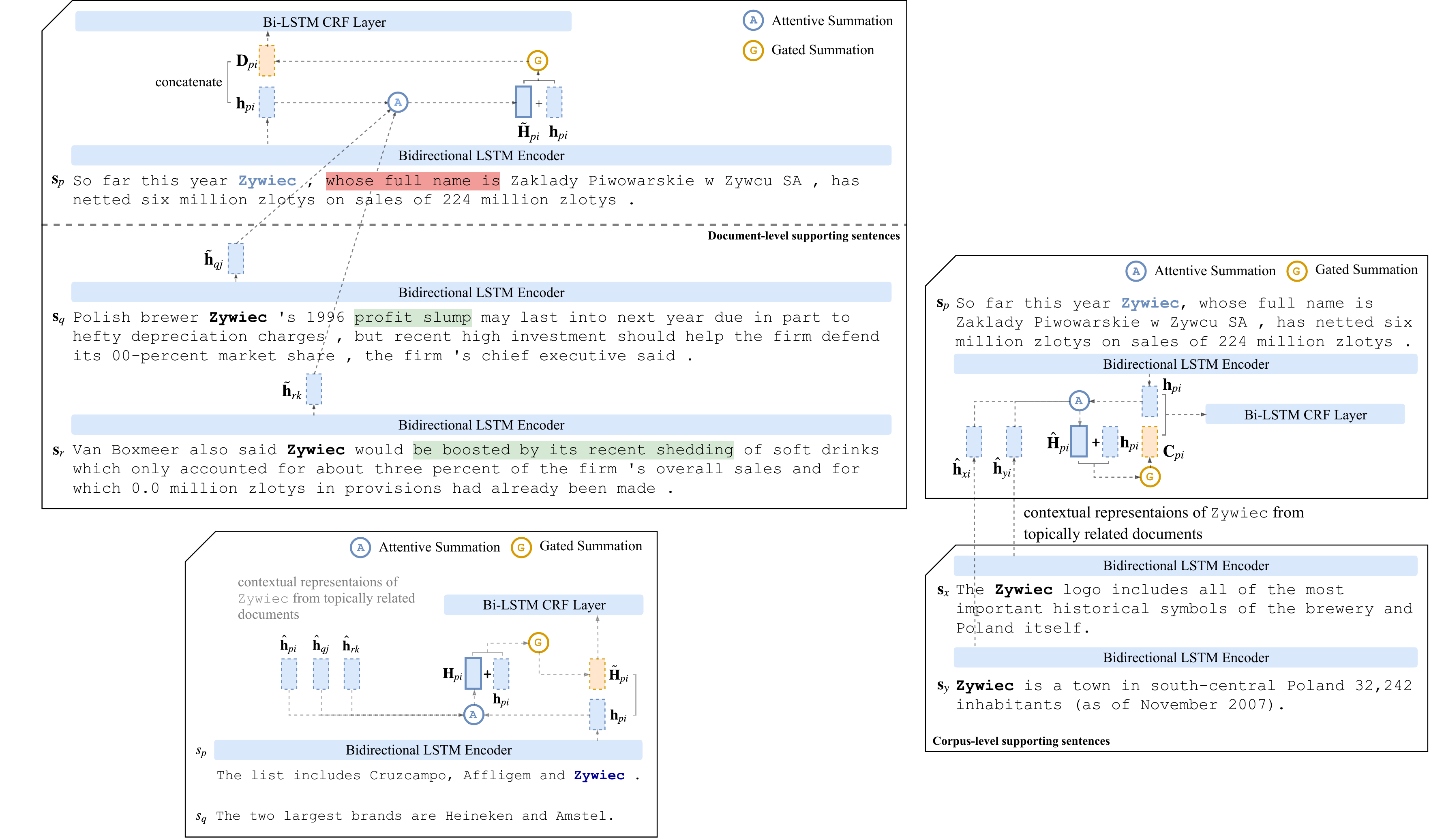}
\vspace{+0.2em}  
   \caption{Document-level Attention Architecture. (Within-sequence context in red incorrectly indicates the name as PER, and document-level context in green correctly indicates the name as ORG.)
   }
   \label{fig:doc_attention}
\end{figure*} 


Many entity mentions are tagged as multiple types by the baseline approach within the same document due to ambiguous contexts (14.43\% of the errors in English, 18.55\% in Dutch, and 17.81\% in German). This type of error is challenging to address as most of the current neural network based approaches focus on evidence within the sentence when making decisions. In cases where a sentence is short or highly ambiguous, the model may either fail to identify names due to insufficient information or make wrong decisions by using noisy context. In contrast, a human in this situation may seek additional evidence from other sentences within the same document to improve judgments.

In Figure~\ref{fig:global_context_example}, the baseline model mistakenly tags ``\texttt{Zywiec}'' as PER due to the ambiguous context ``\texttt{whose full name is}...'', which frequently appears around a person's name. However, contexts from other sentences in the same document containing ``\texttt{Zywiec}'' ({\sl e.g.,} $s_{q}$ and $s_{r}$ in Figure~\ref{fig:doc_attention}), such as ``\texttt{'s 1996 profit}...'' and ``\texttt{would be boosted by its recent shedding}...'', indicate that ``\texttt{Zywiec}'' ought to be tagged as ORG. Thus, we incorporate the document-level supporting evidence with the following attention mechanism~\cite{bahdanau2015attn}.
Formally, given a document $D=\{s_{1}, s_{2}, ...\}$, where $s_{i}=\{w_{i1}, w_{i2}, ...\}$ is a sequence of words, we apply a Bi-LSTM to each word in $s_{i}$, generating local contextual representations $h_{i}=\{\textbf{h}_{i1}, \textbf{h}_{i2}, ...\}$. Next, for each $w_{ij}$, we retrieve the sentences in the document that contain $w_{ij}$ ({\sl e.g.,} $s_{q}$ and $s_{r}$ in Figure~\ref{fig:doc_attention}) and select the local contextual representations of $w_{ij}$ from these sentences as supporting evidence, $\tilde{h}_{ij}=\{\tilde{\textbf{h}}^{1}_{ij}, \tilde{\textbf{h}}^{2}_{ij}, ...\}$ ({\sl e.g.,} $\tilde{\textbf{h}}_{qj}$ and $\tilde{\textbf{h}}_{rk}$ in Figure~\ref{fig:doc_attention}), where $h_{ij}$ and $\tilde{h}_{ij}$ are obtained with the same Bi-LSTM. Since each representation in the supporting evidence is not equally valuable to the final prediction, we apply an attention mechanism to weight the contextual representations of the supporting evidence:
\begin{displaymath}
\setlength{\abovedisplayskip}{4pt}
\setlength{\belowdisplayskip}{4pt}
\resizebox{.40 \textwidth}{!}{
$e^{k}_{ij} = \textbf{v}^{\top}\tanh{\left(W_h \textbf{h}_{ij} + W_{\tilde{h}} \tilde{\textbf{h}}^{k}_{ij} + \textbf{b}_e\right)}$
}\ ,
\label{eq:attention1}
\end{displaymath}
\begin{displaymath}
\setlength{\abovedisplayskip}{4pt}
\setlength{\belowdisplayskip}{4pt}
\alpha^{k}_{ij}= \text{Softmax}\left(e^{k}_{ij}\right)\ ,
\label{eq:attention2}
\end{displaymath}
where $\textbf{h}_{ij}$ is the local contextual representation of word $j$ in sentence $s_i$ and $\tilde{\textbf{h}}^{k}_{ij}$ is the $k$-th supporting contextual representation. $W_h$, $W_{\tilde{h}}$ and $\textbf{b}_e$ are learned parameters. We compute the weighted average of the supporting representations by
\begin{displaymath}
\setlength{\abovedisplayskip}{4pt}
\setlength{\belowdisplayskip}{4pt}
\tilde{\textbf{H}}_{ij}=\sum_{k=1}\alpha^k_{ij} \tilde{\textbf{h}}^{k}_{ij}\ ,
\label{eq:attention3}
\end{displaymath}
where $\tilde{\textbf{H}}_{ij}$ denotes the contextual representation of  the supporting evidence for $w_{ij}$.

For each word $w_{ij}$, its supporting evidence representation, $\tilde{\textbf{H}}_{ij}$, provides a summary of the other contexts where the word appears. Though this evidence is valuable to the prediction process, we must mitigate the influence of the supporting evidence since the prediction should still be made primarily based on the query context. Therefore, we apply a gating mechanism to constrain this influence and enable the model to decide the amount of the supporting evidence that should be incorporated in the prediction process, which is given by
\begin{itemize}
[leftmargin=*,topsep=6pt,itemsep=6pt,parsep=6pt]
\item[] $\textbf{r}_{ij} = \sigma(W_{\tilde{H},r}\tilde{\textbf{H}}_{ij} + W_{h,r}\textbf{h}_{ij} + \textbf{b}_r)$\ ,
\item[] $\textbf{z}_{ij} = \sigma(W_{\tilde{H},z}\tilde{\textbf{H}}_{ij} + W_{h,z}\textbf{h}_{ij} + \textbf{b}_z)$\ ,
\item[] $\textbf{g}_{ij} = \tanh(W_{h,g}\textbf{h}_{ij} + \textbf{z}_{ij} \odot (W_{\tilde{H},g}\tilde{\textbf{H}}_{ij} + \textbf{b}_{g}))$\ ,
\item[] $\textbf{D}_{ij} = \textbf{r}_{ij} \odot \textbf{h}_{ij} + (1 - \textbf{r}_{ij}) \odot \textbf{g}_{ij}$\ ,
\end{itemize}
where all $W$, $\textbf{b}$ are learned parameters and $\textbf{D}_{ij}$ is the gated supporting evidence representation for $w_{ij}$. 



%% file: 3.3corpus_attention.tex
\subsection{Topic-aware Corpus-level Attention}
The document-level attention fails to generate supporting evidence when the name appears only once in a single document. In such situations, we analogously select supporting sentences from the entire corpus. Unfortunately, different from the sentences that are naturally topically relevant within the same documents, the supporting sentences from the other documents may be about distinct topics or scenarios, and identical phrases may refer to various entities with different types, as in the example in Figure~\ref{fig:global_context_example}.
To narrow down the search scope from the entire corpus and avoid unnecessary noise, we introduce a topic-aware corpus-level attention which clusters the documents by topic and carefully selects topically related sentences to use as supporting evidence.

\begin{figure}[h!t]
\centering
\includegraphics[width=0.48\textwidth]{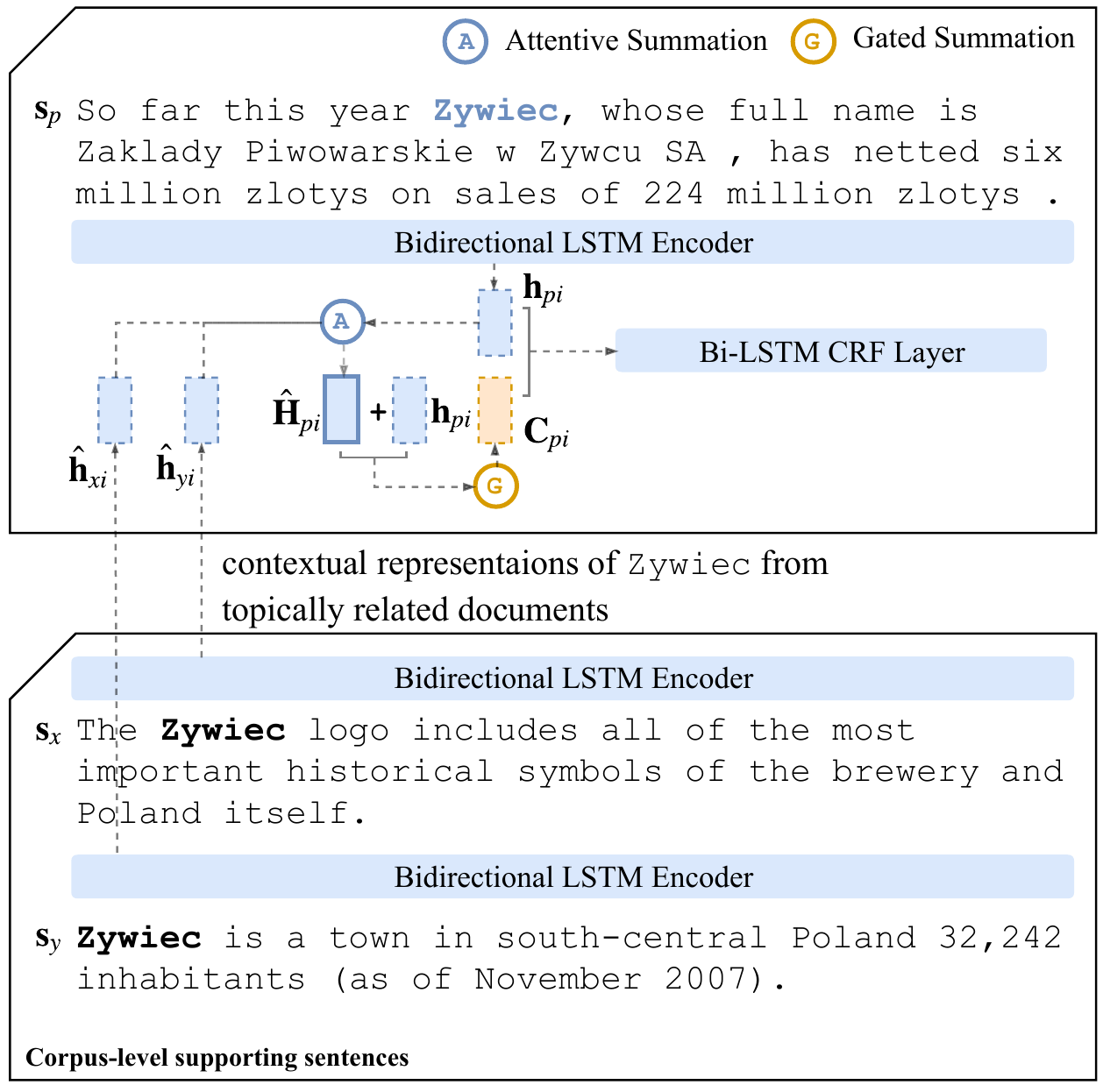}
   \caption{Corpus-level Attention Architecture.
   }
   \label{fig:corpus_attention}
\end{figure}

We first apply Latent Dirichlet allocation (LDA)~\cite{blei2003latent} to model the topic distribution of each document and separate the documents into $N$ clusters based on their topic distributions.\footnote{$N=20$ in our experiments.} As in Figure~\ref{fig:corpus_attention}, we retrieve supporting sentences for each word, such as ``\texttt{Zywiec}'', from the topically related documents and employ another attention mechanism~\cite{bahdanau2015attn} to the supporting contextual representations, $\hat{h}_{ij}=\{\hat{\textbf{h}}^{1}_{ij}, \hat{\textbf{h}}^{2}_{ij}, ...\}$ ({\sl e.g.,} $\tilde{\textbf{h}}_{xi}$ and $\tilde{\textbf{h}}_{yi}$ in Figure~\ref{fig:corpus_attention}). This yields a weighted contextual representation of the corpus-level supporting evidence, $\hat{\textbf{H}}_{ij}$, for each $w_{ij}$, which is similar to the document-level supporting evidence representation, $\tilde{\textbf{H}}_{ij}$, described in section~\ref{section:doc-lvl}. We use another gating mechanism to combine $\hat{\textbf{H}}_{ij}$ and the local contextual representation, $\textbf{h}_{ij}$, to obtain the corpus-level gated supporting evidence representation, $\textbf{C}_{ij}$, for each $w_{ij}$.

\graphicspath { {./fig/} }

%% file: 3.4prediction.tex
\subsection{Tag Prediction}
For each word $w_{ij}$ of sentence $s_i$, we concatenate its local contextual representation $\textbf{h}_{ij}$, document-level gated supporting evidence representation $\textbf{D}_{ij}$, and corpus-level gated supporting evidence representation $\textbf{C}_{ij}$ to obtain its final representation. This representation is fed to another Bi-LSTM to further encode the supporting evidence and local contextual features into an unified representation, which is given as input to an affine-CRF layer for label prediction.

%% file: 4experiments.tex
\section{Experiments}

\input{4.1dataset}
\input{4.2setup}

\input{4.3quantitative}
\input{4.4qualitative}
\input{4.5errors}

%% file: 4.1dataset.tex
\subsection{Dataset}\label{subsec:data}

We evaluate our methods on the CoNLL-2002 and CoNLL-2003 name tagging datasets~\cite{tjong2003introduction}. The CoNLL-2002 dataset contains name tagging annotations for Dutch (NLD) and Spanish (ESP), while the CoNLL-2003 dataset contains annotations for English (ENG) and German (DEU). Both datasets have four pre-defined name types: person (PER), organization (ORG), location (LOC) and miscellaneous (MISC).\footnote{The miscellaneous category consists of names that do not belong to the other three categories.}

\begin{table}[!hbt]
\setlength\tabcolsep{3pt}
\setlength\extrarowheight{1pt}
\small
\begin{tabularx}{\linewidth}{|>{\hsize=.5\hsize}X|>{\hsize=1.3\hsize}X|>{\hsize=1.1\hsize}X|>{\hsize=1.1\hsize}X|}
\hline
\textbf{Code} & \textbf{Train} & \textbf{Dev.} & \textbf{Test}  \\
\hline
NLD &  202,931 (13,344) & 37,761 (2,616) & 68,994 (3,941) \\
ESP & 264,715 (18,797) & 52,923 (4,351) & 51,533 (3,558) \\
ENG & 204,567 (23,499) & 51,578 (5,942) & 46,666 (5,648)\\
DEU &  207,484 (11,651) & 51,645 (4,669) & 52,098 (3,602) \\
\hline
\end{tabularx}
\label{tab:data}
\caption{\# of tokens in name tagging datasets statistics. \# of names is given in parentheses.}
\end{table}

We select at most four document-level supporting sentences and five corpus-level supporting sentences.\footnote{Both numbers are tuned from 1 to 10 and selected when the model performs best on the development set.} Since the document-level attention method requires input from each individual document, we do not evaluate it on the CoNLL-2002 Spanish dataset which lacks document delimiters. We still evaluate the corpus-level attention on the Spanish dataset by randomly splitting the dataset into documents (30 sentences per document). Although randomly splitting the sentences does not yield perfect topic modeling clusters, experiments show the corpus-level attention still outperforms the baseline (Section~\ref{sec:results_analysis}).

%% file: 4.2setup.tex
\subsection{Experimental Setup}
For word representations, we use 100-dimensional pre-trained word embeddings and 25-dimensional randomly initialized character embeddings. We train word embeddings using the word2vec package.\footnote{https://github.com/tmikolov/word2vec} English embeddings are trained on the English Giga-word version 4, which is the same corpus used in~\cite{lample2016neural}. Dutch, Spanish, and German embeddings are trained on corresponding Wikipedia articles (2017-12-20 dumps). Word embeddings are fine-tuned during training.


Table~\ref{tab:parameters} shows our hyper-parameters. For each model with an attention, since the Bi-LSTM encoder must encode the local, document-level, and/or corpus-level contexts, we pre-train a Bi-LSTM CRF model for 50 epochs, add our document-level attention and/or corpus-level attention, and then fine-tune the augmented model. Additionally,~\citet{reimers2017optimal} report that neural models produce different results even with same hyper-parameters due to the variances in parameter initialization. Therefore, we run each model ten times and report the mean as well as the maximum F1 scores.



\begin{table}
\small
\centering
\begin{tabular}{l|l}
\hline
\textbf{Hyper-parameter} & \textbf{Value}  \\
\hline
CharCNN Filter Number & 25 \\
CharCNN Filter Widths & [2, 3, 4] \\
Lower Bi-LSTM Hidden Size &  100 \\
Lower Bi-LSTM Dropout Rate & 0.5 \\
Upper Bi-LSTM Hidden Size & 100 \\
Learning Rate & 0.005 \\
Batch Size & N/A$^*$ \\
Optimizer & SGD~\cite{bottou2010large} \\
\hline
\end{tabular}
\raggedright $*$~Each batch is a document. The batch size varies as the different document length.
\caption{Hyper-parameters.}
\label{tab:parameters}
\end{table}

%% file: 4.3quantitative.tex
\subsection{Performance Comparison}\label{sec:results_analysis}

\begin{table}[h!t]
\centering
\small
\begin{tabular}{l|l|c|l}
\hline
\textbf{Code} & \textbf{Model} & \multicolumn{2}{c}{\textbf{F1 (\%)}} \\
\hline
 & \cite{gillick2015multilingual} & reported & $82.84$ \\
& \cite{lample2016neural} & reported & $81.74$ \\
 \multirow{7}{*}{NLD} & \cite{yang2017transfer} & reported & $85.19$ \\
\cline{2-4}
 & \multirow{2}{*}{Our Baseline} & mean & $85.43$  \\
 \cline{3-4}
& & max & $85.80$ \\
\cline{2-4}
& \multirow{2}{*}{Doc-lvl Attention} & mean & $86.82$ \\
\cline{3-4}
& & max & $87.05$ \\
\cline{2-4}
 & \multirow{2}{*}{Corpus-lvl Attention} & mean & $86.41$ \\
 \cline{3-4}
& & max & $86.88$ \\
\cline{2-4}
 & \multirow{3}{*}{Both} & mean & $87.14$ \\
 \cline{3-4}
& & max & $\textbf{87.40}$ \\
 \cline{3-4}
& & $\Delta$ & $\textbf{+1.60}$ \\
\hhline{====}
\multirow{7}{*}{ESP} & \cite{gillick2015multilingual} & reported & $82.95$ \\
& \cite{lample2016neural} & reported & $85.75$ \\
&  \cite{yang2017transfer} & reported & $85.77$ \\
\cline{2-4}
 & \multirow{2}{*}{Our Baseline} & mean & $85.33$ \\
 \cline{3-4}
& & max & $85.51$ \\
\cline{2-4}
 & \multirow{2}{*}{Corpus-lvl Attention} & mean & $85.77$ \\
 \cline{3-4}
& & max & $\textbf{86.01}$ \\
 \cline{3-4}
& & $\Delta$ & $\textbf{+0.50}$ \\
\hhline{====}
& \cite{luo2015joint} & reported & $91.20$ \\
\multirow{10}{*}{ENG} & \cite{lample2016neural} & reported & $90.94$ \\
& \cite{ma-hovy:2016:P16-1} & reported & $91.21$ \\
& \cite{liu2017empower} & reported & $91.35$ \\
& \cite{peters2017semi} & reported & $91.93$ \\
& \cite{peters2018deep} & reported & $\textbf{92.22}$ \\
\cline{2-4}
 & \multirow{2}{*}{Our Baseline} & mean & $90.97$ \\
 \cline{3-4}
& & max & $91.23$ \\
\cline{2-4}
& \multirow{2}{*}{Doc-lvl Attention} & mean & $91.43$ \\
\cline{3-4}
& & max & $91.60$ \\
\cline{2-4}
 & \multirow{2}{*}{Corpus-lvl Attention} & mean & $91.41$ \\
 \cline{3-4}
& & max & $91.71$ \\
\cline{2-4}
 & \multirow{3}{*}{Both} & mean & $91.64$\\
 \cline{3-4}
& & max & $\textbf{91.81}$ \\
 \cline{3-4}
& & $\Delta$ & $\textbf{+0.58}$ \\
\hhline{====}
 & \cite{gillick2015multilingual} & reported & $76.22$ \\
\multirow{7}{*}{DEU} & \cite{lample2016neural} & reported & $78.76$ \\
\cline{2-4}
 & \multirow{2}{*}{Our Baseline} & mean & $78.15$ \\
 \cline{3-4}
& & max & $78.42$ \\
\cline{2-4}
& \multirow{2}{*}{Doc-lvl Attention} & mean & $78.90$ \\
\cline{3-4}
& & max & $79.19$ \\
\cline{2-4}
 & \multirow{2}{*}{Corpus-lvl Attention} & mean & $78.53$ \\
 \cline{3-4}
& & max & $78.88$ \\
\cline{2-4}
 & \multirow{3}{*}{Both} & mean & $78.83$ \\
 \cline{3-4}
& & max & $\textbf{79.21}$ \\ 
\cline{3-4}
& & $\Delta$ & $\textbf{+0.79}$ \\
\hline
\end{tabular}
\caption{Performance of our methods versus the baseline and state-of-the-art models.}
\label{tab:results}
\end{table}

\graphicspath { {./fig/} }
\begin{figure*}[h!t]
\begin{subfigure}{.5\textwidth}
  \centering
  \includegraphics[width=1\linewidth]{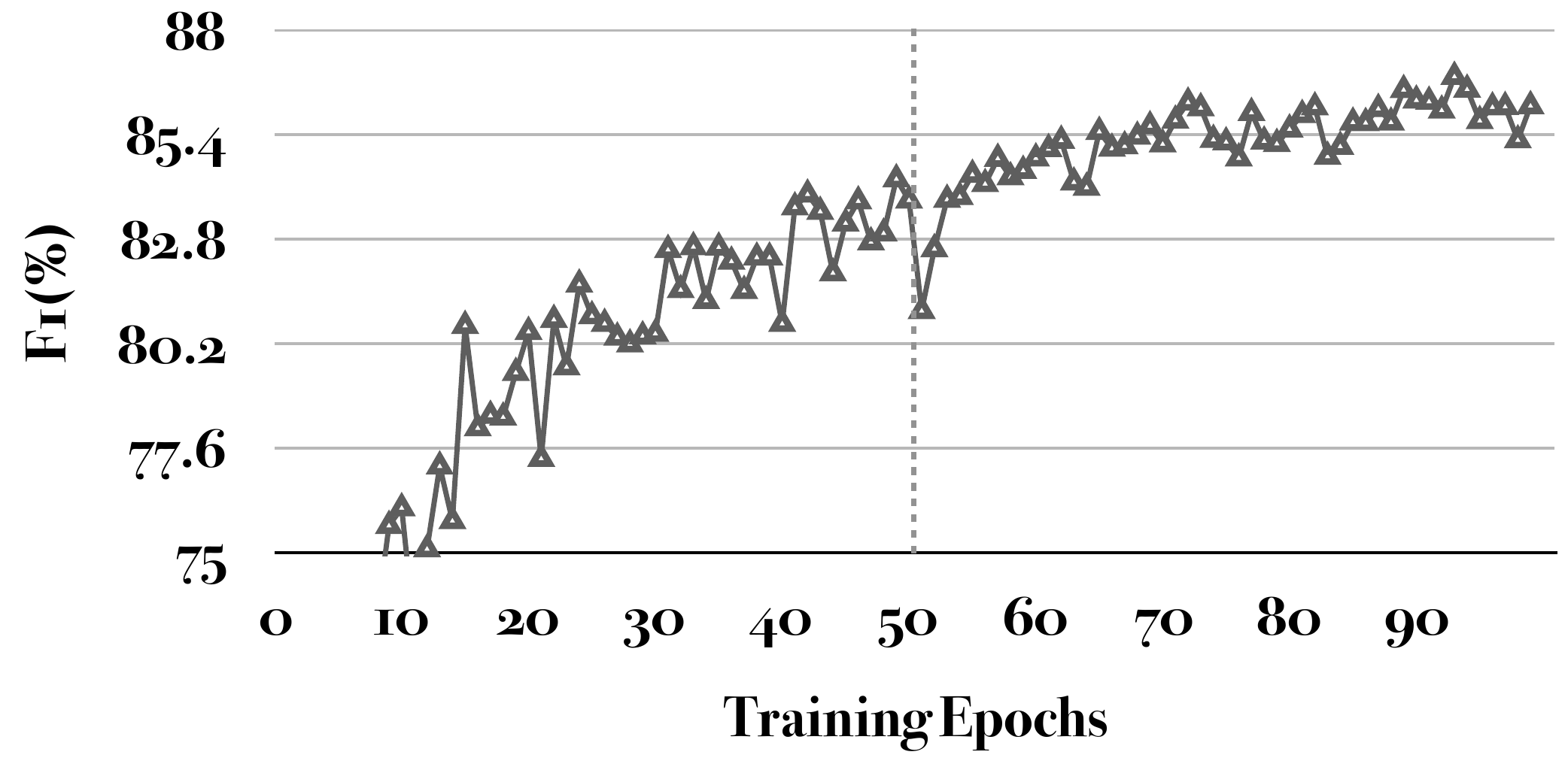}
  \caption{Dutch (F1 scales between 75\%-88\%)}
  \label{fig:nld_curve}
\end{subfigure}
\begin{subfigure}{.5\textwidth}
  \centering
  \includegraphics[width=1\linewidth]{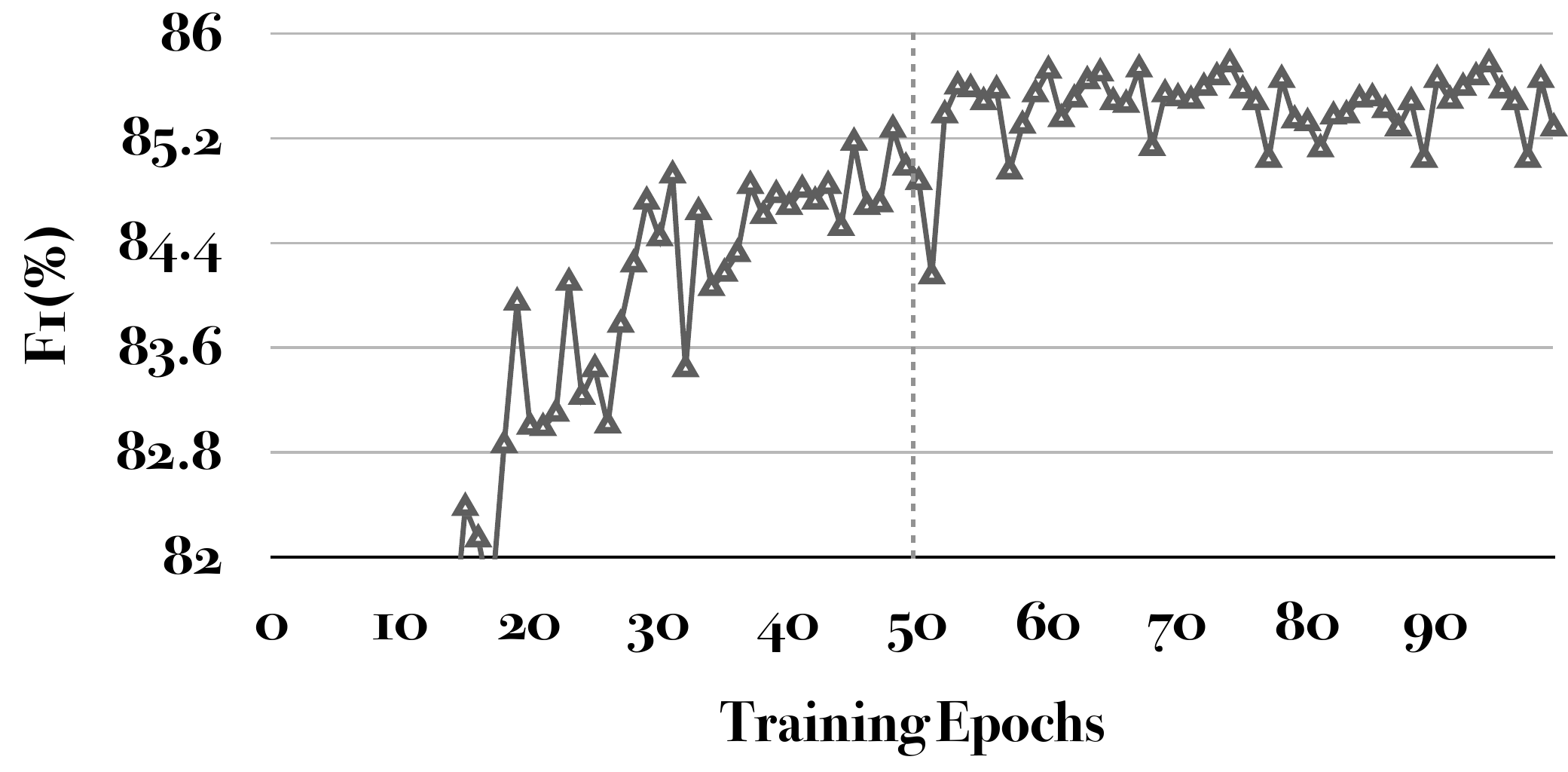}
  \caption{Spanish (F1 scales between 82\%-86\%)}
  \label{fig:esp_curve}
\end{subfigure}
\begin{subfigure}{.5\textwidth}
  \centering
  \includegraphics[width=1\linewidth]{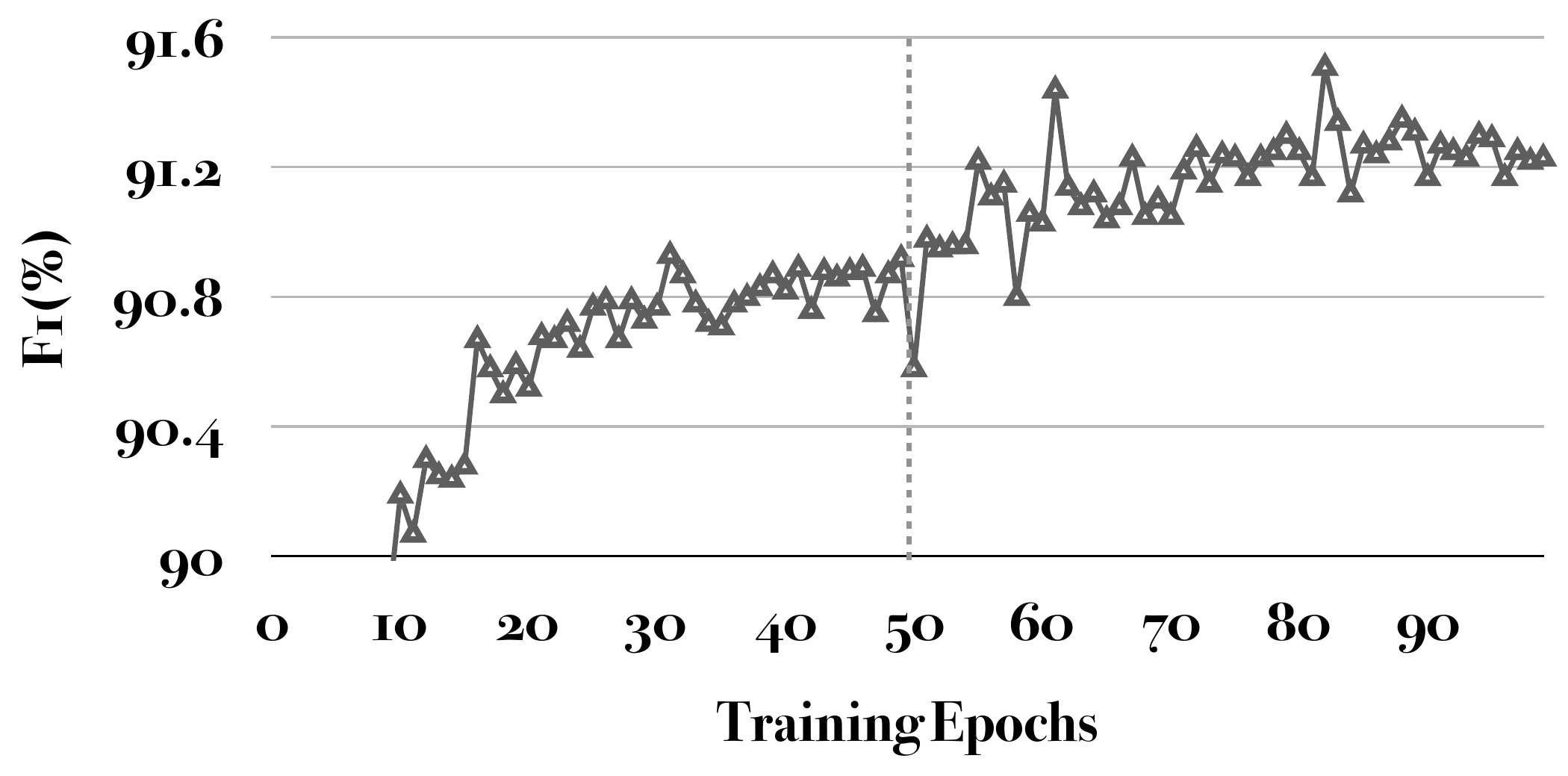}
  \caption{English (F1 scales between 90\%-91.6\%)}
  \label{fig:eng_curve}
\end{subfigure}
\begin{subfigure}{.5\textwidth}
  \centering
  \includegraphics[width=1\linewidth]{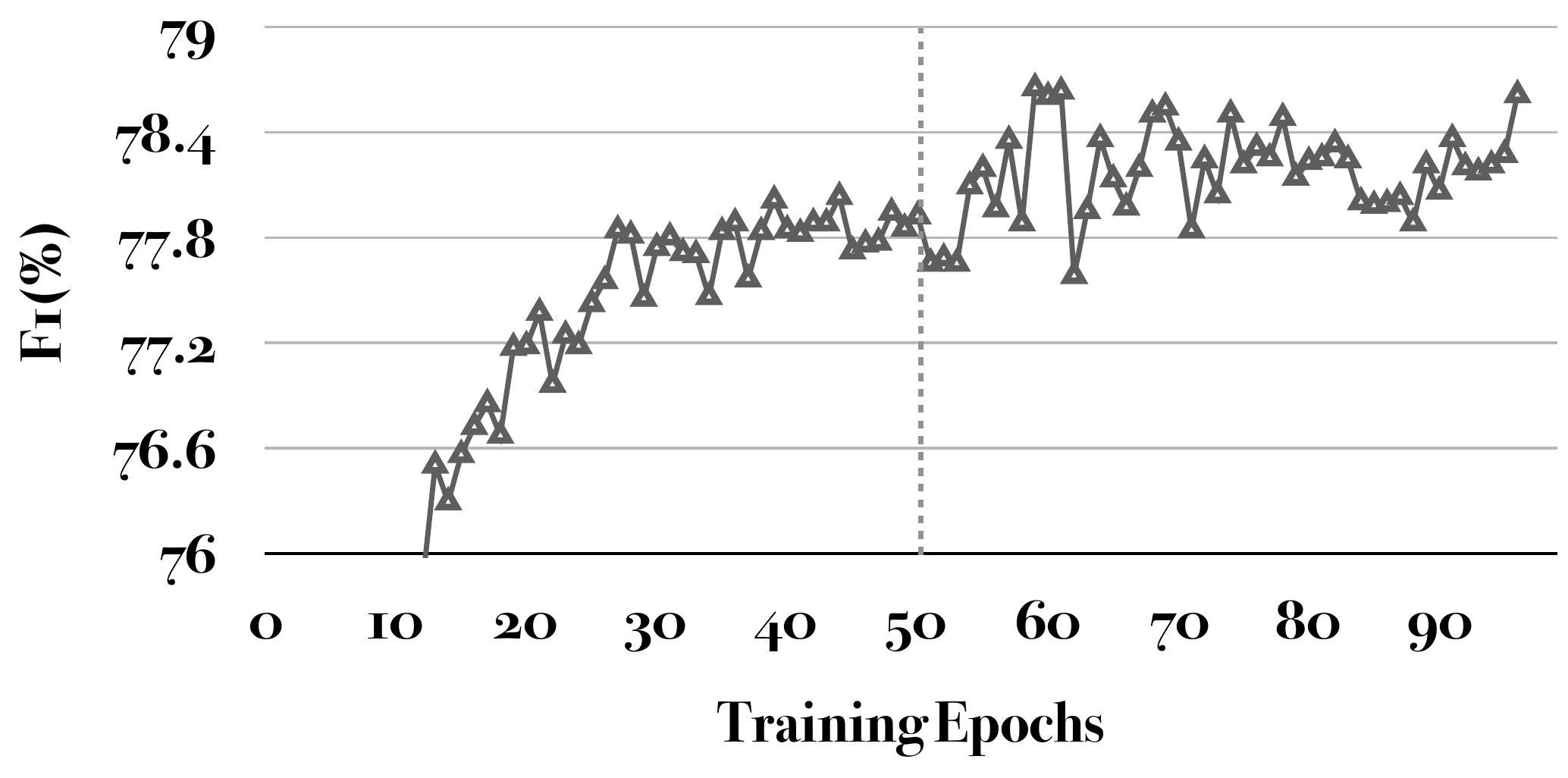}
  \caption{German (F1 scales between 76\%-79\%)}
  \label{fig:deu_curve}
\end{subfigure}
\caption{Average F1 score for each epoch of the ten runs of our model with both document-level and corpus-level attentions. Epochs 1-50 are the pre-training phase and 51-100 are the fine-tuning phase.}
\label{fig:learning_curves}
\end{figure*}

We compare our methods to three categories of baseline name tagging methods:
\begin{itemize}[leftmargin=*,topsep=6pt,itemsep=6pt,parsep=6pt]
\item \textbf{Vanilla Name Tagging}
Without any additional resources and supervision, the current state-of-the-art name tagging model is the Bi-LSTM-CRF network reported by~\citet{lample2016neural} and~\citet{ma-hovy:2016:P16-1}, whose difference lies in using a LSTM or CNN to encode characters. Our methods fall in this category.
\item \textbf{Multi-task Learning}
\citet{luo2015joint,yang2017transfer} apply multi-task learning to boost name tagging performance by introducing additional annotations from related tasks such as entity linking and part-of-speech tagging.
\item \textbf{Join-learning with Language Model}
\citet{peters2017semi,liu2017empower,peters2018deep} leverage a  pre-trained language model on a large external corpus to enhance the semantic representations of words in the local corpus. \citet{peters2018deep} achieve a high score on the CoNLL-2003 English dataset using a giant language model pre-trained on a 1 Billion Word Benchmark~\cite{chelba2013one}.
\end{itemize}

Table~\ref{tab:results} presents the performance comparison among the baselines, the aforementioned state-of-the-art methods, and our proposed methods.
Adding only the document-level attention offers a F1 gain of between $0.37\%$ and $1.25\%$ on Dutch, English, and German. Similarly, the addition of the corpus-level attention yields a F1 gain between $0.46\%$ to $1.08\%$ across all four languages. The model with both attentions outperforms our baseline method by $1.60\%$, $0.56\%$, and $0.79\%$ on Dutch, English, and German, respectively. Using a paired t-test between our proposed model and the baselines on 10 randomly sampled subsets, we find that the improvements are statistically significant ($p \leq 0.015$) for all settings and all languages.


By incorporating the document-level and corpus-level attentions, we achieve state-of-the-art performance on the Dutch (NLD), Spanish (ESP) and German (DEU) datasets. For English, our methods outperform the state-of-the-art methods in the ``Vanilla Name Tagging'' category. Since the document-level and corpus-level attentions introduce redundant and topically related information, our models are compatible with the language model enhanced approaches. It is interesting to explore the integration of these two methods, but we leave this to future explorations.

Figure~\ref{fig:learning_curves} presents, for each language, the learning curves of the full models ({\sl i.e.,} with both document-level and corpus-level attentions). The learning curve is computed by averaging the F1 scores of the ten runs at each epoch. We first pre-train a baseline Bi-LSTM CRF model from epoch 1 to 50. Then, starting at epoch 51, we incorporate the document-level and corpus-level attentions to fine-tune the entire model. As shown in Figure~\ref{fig:learning_curves}, when adding the attentions at epoch 51, the F1 score drops significantly as new parameters are introduced to the model. The model gradually adapts to the new information, the F1 score rises, and the full model eventually outperforms the pre-trained model. The learning curves strongly prove the effectiveness of our proposed methods.

We also compare our approach with a simple rule-based propagation method, where we use token-level majority voting to make labels consistent on document-level and corpus-level. The score of document-level propagation on English is 90.21\% (F1), and the corpus-level propagation is 89.02\% which are both lower than the BiLSTM-CRF baseline 90.97\%.

%% file: 4.4qualitative.tex
\subsection{Qualitative Analysis}

\begin{figure*}[h!t]
\centering
\includegraphics[width=0.99\textwidth]{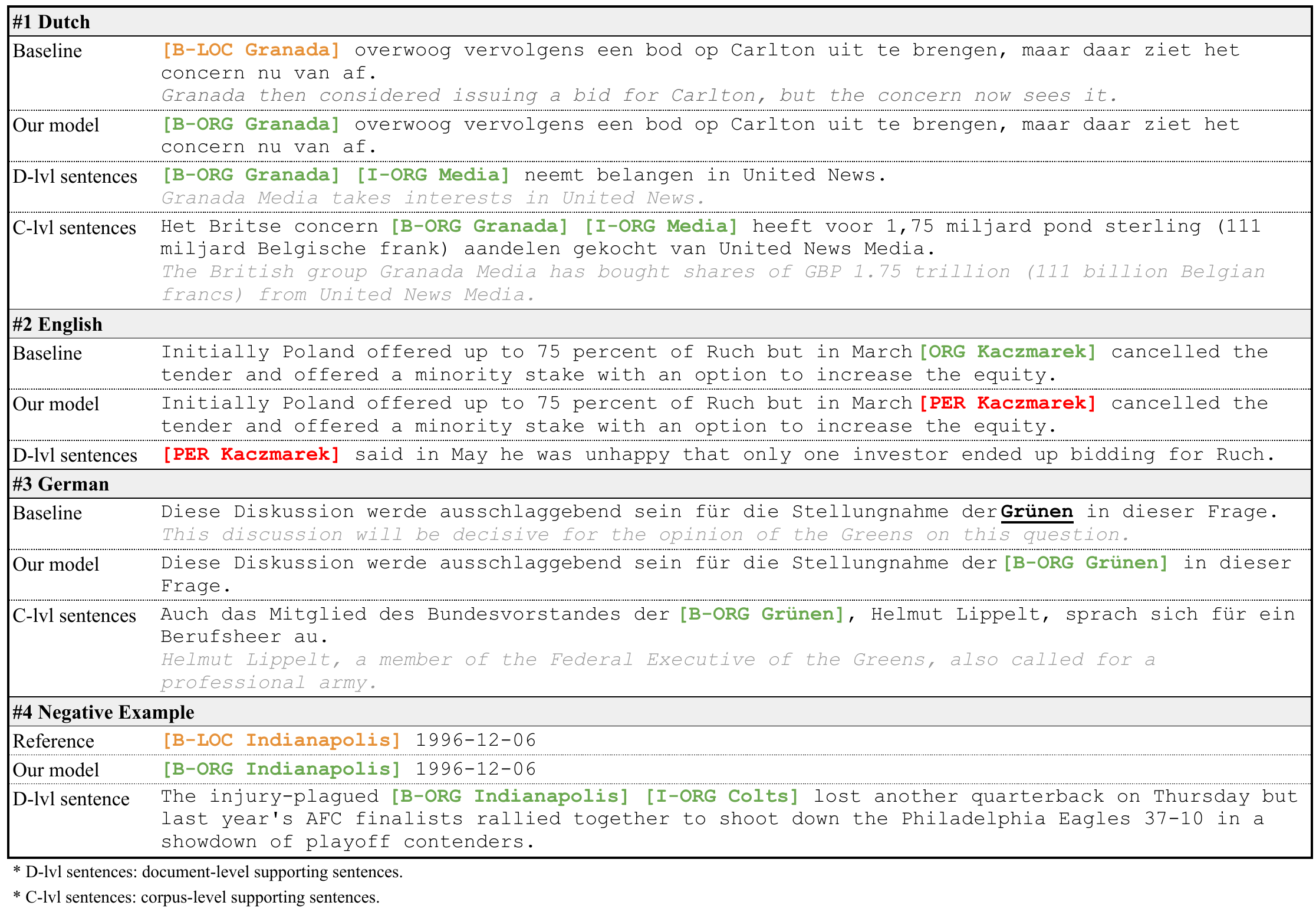}
\caption{Comparison of name tagging results between the baseline and our methods.}
\label{fig:qualitative_analysis}
\end{figure*}

Table~\ref{fig:qualitative_analysis} compares the name tagging results from the baseline model and our best models. All examples are selected from the development set.

In the Dutch example, ``\texttt{Granada}'' is the name of a city in Spain, but also the short name of ``\texttt{Granada Media}''. Without ORG related context, ``\texttt{Granada}'' is mistakenly tagged as LOC by the baseline model. However, the document-level and corpus-level supporting evidence retrieved by our method contains the ORG name ``\texttt{Granada Media}'', which strongly indicates ``\texttt{Granada}'' to be an ORG in the query sentence. By adding the document-level and corpus-level attentions, our model successfully tags ``\texttt{Granada}'' as ORG.

In example 2, the OOV word ``\texttt{Kaczmarek}'' is tagged as ORG in the baseline output. In the retrieved document-level supporting sentences, PER related contextual information, such as the pronoun ``\texttt{he}'', indicates ``\texttt{Kaczmarek}'' to be a PER. Our model correctly tags ``\texttt{Kaczmarek}'' as PER with the document-level attention.

In the German example, ``\texttt{Grünen}'' (Greens) is an OOV word in the training set. The character embedding captures the semantic meaning of the stem ``\texttt{Grün}'' (Green) which is a common non-name word, so the baseline model tags ``\texttt{Grünen}'' as O (outside of a name). In contrast, our model makes the correct prediction by incorporating the corpus-level attention because in the related sentence from the corpus ``\texttt{Bundesvorstandes der Grünen}'' (Federal Executive of the Greens) indicates ``\texttt{Grünen}'' to be a company name.

\graphicspath { {./fig/} }

%% file: 4.5errors.tex
\subsection{Remaining Challenges}
By investigating the remaining errors, most of the named entity type inconsistency errors are eliminated, however, a few new errors are introduced due to the model propagating labels from negative instances to positive ones. Figure~\ref{fig:qualitative_analysis} presents a negative example, where our model, being influenced by the prediction ``\texttt{[B-ORG Indianapolis]}'' in the supporting sentence, incorrectly predicts ``\texttt{Indianapolis}'' as ORG in the query sentence. 
A potential solution is to apply sentence classification~\cite{kim2014convolutional,ji2017neural} to the documents, divide the document into fine-grained clusters of sentences, and select supporting sentences within the same cluster.

In morphologically rich languages, words may have many variants. When retrieving supporting evidence, our exact query word match criterion misses potentially useful supporting sentences that contain variants of the word.
Normalization and morphological analysis can be applied in this case to help fetch supporting sentences.




%% file: 2related.tex
\section{Related Work}
Name tagging methods based on sequence labeling have been extensively studied recently.~\citet{huang2015bidirectional} and~\citet{lample2016neural} proposed a neural architecture consisting of a bi-directional long short-term memory network (Bi-LSTM) encoder and a conditional random field (CRF) output layer (Bi-LSTM CRF). This architecture has been widely explored and demonstrated to be effective for sequence labeling tasks. Efforts incorporated character level compositional word embeddings, language modeling, and CRF re-ranking into the Bi-LSTM CRF architecture which improved the performance~\cite{ma2016end,liu2017empower,sato2017segment,peters2017semi,peters2018deep}. Similar to these studies, our approach is also based on a Bi-LSTM CRF architecture. However, considering the limited contexts within each individual sequence, we design two attention mechanisms to further incorporate topically related contextual information on both the document-level and corpus-level.

There have been efforts in other areas of information extraction to exploit features beyond individual sequences. Early attempts~\cite{mikheev1998description,mikheev2000document} on MUC-7 name tagging dataset used document centered approaches. A number of approaches explored document-level features ({\sl e.g.,} temporal and co-occurrence patterns) for event extraction~\cite{chambers2008jointly,ji2008refining,liao2010using,do2012joint,mcclosky2012learning,berant2014modeling,yang2016joint}. Other approaches leveraged features from external resources ({\sl e.g.,} Wiktionary or FrameNet) for low resource name tagging and event extraction~\cite{li2013joint,huang2016liberal,liu2016leveraging,zhang2016name,cotterell2017low,zhang2017embracing,huang2018multi}.~\citet{yaghoobzadeh2016corpus} aggregated corpus-level contextual information of each entity to predict its type and~\citet{narasimhan2016improving} incorporated contexts from external information sources ({\sl e.g.,} the documents that contain the desired information) to resolve ambiguities. Compared with these studies, our work incorporates both document-level and corpus-level contextual information with attention mechanisms, which is a more advanced and efficient way to capture meaningful additional features. Additionally, our model is able to learn how to regulate the influence of the information outside the local context using gating mechanisms.

%% file: 5conclusions.tex
\section{Conclusions and Future Work}
We propose document-level and corpus-level attentions for name tagging. The document-level attention retrieves additional supporting evidence from other sentences within the document to enhance the local contextual information of the query word. When the query word is unique in the document, the corpus-level attention searches for topically related sentences in the corpus. Both attentions dynamically weight the retrieved contextual information and emphasize the information most relevant to the query context. We present gating mechanisms that allow the model to regulate the influence of the supporting evidence on the predictions. Experiments demonstrate the effectiveness of our approach, which achieves state-of-the-art results on benchmark datasets.

We plan to apply our method to other tasks, such as event extraction, and explore integrating language modeling into this architecture to further boost name tagging performance.